# SPACE ROBOTICS

## Part 2: Space-based Manipulators

**Alex Ellery**
Surrey Space Centre, University of Surrey, Guildford, Surrey, United Kingdom
Contact: a.ellery@eim.surrey.ac.uk

### 1. Introduction

Few engineered systems are expected to survive and function for more than a few years up to a decade or more without human intervention for servicing, maintenance or upgrading. Spacecraft are one of the few, engineered long-life products of human society that are denied such service and maintenance as part of their operational lifecycle. Simplistic notions of inaccessibility are no longer tenable as an excuse for this - the technology is in place to realise robotic on-orbit servicing (OOS). The traditional approach to spacecraft reliability has been through emphasis on high reliability, high cost components, and extensive validation and testing which also contribute to the expense of space platforms. The recent worrying trend in increasing on-orbit spacecraft failures has provided considerable support to the failure of this approach (Sullivan, B. & Akin, D., 2001). For any space platform, it is desirable to increase operational availability which requires a mixed approach:

Availability,

$$A = (MTBF/(MTBF+MTTR+MTFS)) \qquad (1)$$

where
MTBF=mean time between failures and reflects reliability
MTTR=mean time to repair and reflects maintainability
MTFS=mean time for supply and reflects logistic capability

**In this second of three short papers, I introduce some of the basic concepts of space robotics with an emphasis on some specific challenging areas of research that are peculiar to the application of robotics to space infrastructure development. The style of these short papers is pedagogical and the concepts in this paper are developed from fundamental manipulator robotics. This second paper considers the application of space manipulators to on-orbit servicing (OOS), an application which has considerable commercial application. I provide some background to the notion of robotic on-orbit servicing and explore how manipulator control algorithms may be modified to accommodate space manipulators which operate in the micro-gravity of space.**

Given the failure of reliability alone approaches to maximising spacecraft availability, maintenance of Earth-orbiting spacecraft through on-orbit servicing is essential by reducing MTTR and MTFS.

Servicing of satellites may be implemented in all major orbits currently inhabited by Earth orbiting satellites. There are a number of families of orbits used by spacecraft today. Low Earth orbit (LEO) capped by the lowest point of the inner van Allen radiation belt at 2,000 km altitude is utilised by human missions and Earth observation missions (at polar inclinations). Medium Earth orbit (MEO) resides between the inner and outer Van Allen radiation belts centred around 10,000 km altitude and is ideal for mobile satellite constellations, eg. GPS constellation reside in 18,000 km altitude orbits. Most communications satellites reside in geosynchronous equatorial orbits (GEO) at 36,000 km altitude (though many Russian satellites utilise the high inclination Molniya orbits for access to high latitudes). In addition, there are highly elliptical orbits (HEO) that are used for some astronomy missions. For future astronomy missions, the Sun-Earth $L_1$ (for solar observations, eg. SOHO) and $L_2$ (for deep space observations, eg. Microwave Anisotropy Probe, Next Generation Space Telescope, Terrestrial Planet Finder) Lagrange points are expected to be the orbits of choice. On-orbit servicing has considerable potential for commercial applications in providing a space-based infrastructure (Ellery, A., Welch, C., & Curley, A., 2001). It has been suggested that the European Robotic

Arm (ERA) on the ISS might be used to support astronomy missions by upgrading their instruments as an ISS-based servicer manipulator (Ellery, A. 2003).

## 2. On-Orbit Servicing

The Solar Maximum Repair Mission (SMRM) of 1984 was the first demonstration of on-orbit servicing by astronauts in combination with software workarounds uploaded from the ground, and teleoperation of the Shuttle Remote Manipulator System (SRMS) by an astronaut. The Solar Maximum Repair Mission represents a "textbook" case of OOS, involving the exchange of ORU (Orbital Replacement Unit) modules. Although the more complex tasks were performed by astronauts on EVA (extravehicular activity), such servicing operations may potentially be performed by robotic servicers. The repair and servicing of the Hubble Space Telescope and other US astronaut activities have further demonstrated the feasibility of space-based servicing tasks. Indeed, robotic servicing was an instrumental part of the early stages of the ISS programme in which two concepts were proposed to perform these functions - the Flight Telerobotic Servicer (FTS) and the Orbital Maneouvring Vehicle (OMV) - but these were cancelled in the face of budget cuts. NASA's AERCam (Autonomous Extravehicular Robotic CAMera) represents a step in this direction – AERCam is a small 35 cm diameter freeflying sphere comprising a camera for aiding astronaut EVA, thrusters for attitude and translation control, and avionics developed from astronaut MMU (manned maneouvering unit) technology. The addition of robotic manipulators onto a larger spacecraft platform would offer freeflyer servicer capabilities. The sizing of the manipulator would be determined by EVA-equivalence, one example of which is the proposed ESA dextrous robot manipulator system:

1. Seven degrees of freedom (three degrees of freedom at the shoulder, one degree of freedom at the elbow, and three degrees of freedom at the spherical wrist)
2. Three fingered end effector with cylindrical dimensions 10 x 15 cm
3. Control set-point rate of 100 Hz
4. Forward reach of 1m - this requires multiple grappling points on the target as full reachability around most satellites would require a reach of 4.5 - 16 m which is impractical
5. End effector position accuracy of 0.3 mm/0.1º
6. Maximum end effector velocity of 0.1 m/s and 0.1 rad/s
7. Structural displacement compliance of $1 \times 10^6$ N/m and rotational compliance of $5 \times 10^4$ Nm/rad
8. Force/torque exertion of 200 N and 20 Nm respectively
9. Payload capacity of 500 kg in microgravity environment

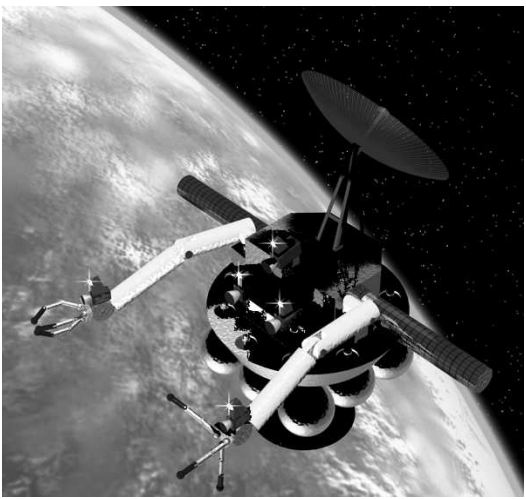
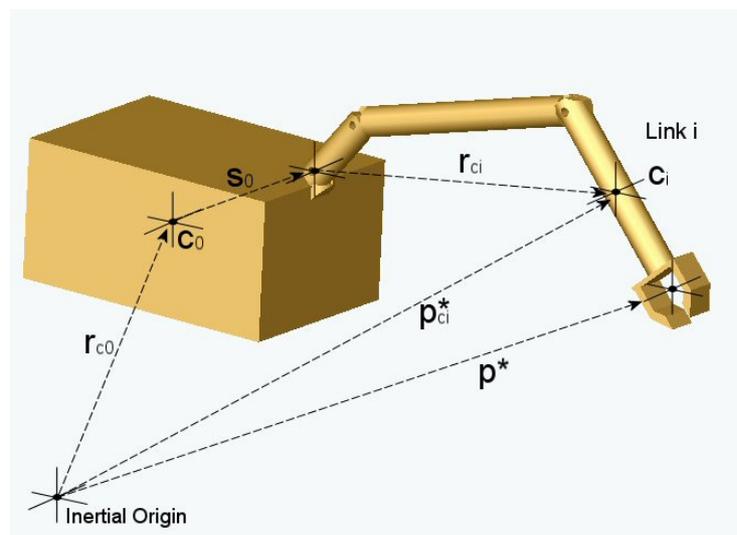

Fig. 1. ATLAS robotic servicer concept kinematics and dynamics

Fig. 2. Single arm version of the robotic servicer (courtesy Praxis Publishers)

The Japanese ETS (Engineering Test Satellite) VII of 1996 has demonstrated many of the basic technologies for robotic on-orbit servicing, including autonomous rendezvous and docking to a target, teleoperated manipulator movement control whilst maintaining stable attitude, and vision/force feedback based peg-in-hole tasks. All robotic servicers will be required to grapple the target spacecraft for retrieval, resupply of consumables, repair, or retrofit. In the US, one of the most advanced servicer concepts is Ranger for which a neutral buoyancy test vehicle has been developed. One proposal for such a robotic servicer is 1.5 tonne ATLAS (Advanced TeleRobotic Actuation System) – see Fig. 1.:

## 3. Control of Space Manipulators

On-orbit servicing robotics is a modern version of an old field that stems back to the origins of science itself – Newton (1642-1729), Euler (1707-1783), d'Alembert (1717-1783), Lagrange (1736-1813), Laplace (1749-1827) and Hamilton (1805-1865) all contributed to the development of mechanics and dynamics. The primary differentiating characteristic of on-orbit servicing robotics from terrestrial robotics is that the robotic servicer operates in microgravity. Whereas the terrestrial manipulator is mounted onto *terra firma*, in space there is no such reaction force and torque cancellation – the motion of the manipulator(s) will generate reaction forces and moments on the spacecraft at the manipulator mounting point(s). Robotic freeflyer manipulators are difficult to control as the spacecraft platform moves in response to the manipulator movements. A free-floating platform no longer permits the use of the base of the manipulator as the inertial coordinate frame of reference. If this effect is not taken into account, the manipulators will overshoot the target that it is attempting to grapple. A similar effect occurs with astronauts in the microgravity environment of space. They undergo changes in psychomotor performance and posture and their limb movements tend to overshoot their targets until the astronaut's brain has adapted to the new microgravity conditions (normally within two to three days). The robotic manipulator control system must similarly compensate for operating in microgravity while implementing the computationally intensive algorithms for trajectory interpolation, inverse kinematics, dynamics, and force/position control of the end effector. We may apply the conservation of momentum to the freeflyer servicer system (assuming a single manipulator for simplicity) in order to apply constraints to solve the problem, which allows us to define the centre of mass of the whole system to lie at the origin of the inertial reference frame - see Fig. 2.:

The position of the manipulator end effector with respect to inertial space may be represented by:

$$p^* = r_{c0} + R_0 s_0 + \sum_{i=1}^{n} R_i l_i \qquad (2)$$

Although a number of global dynamics techniques have been proposed, they suffer from high computational complexity problems (Umetani, Y. & Yoshida, K., 1989). Now, although conservation of linear momentum is integrable to yield constraints on linear position of the end effector, this is not the case for angular momentum conservation, which is not integrable to unique angular constraints as it is a non-holonomic constraint. It is possible to separate out the rotational and translation components of the system to yield a much simpler and more readily implementable set of control algorithms. We employ a dedicated spacecraft attitude control to which a feedforward signal from the robotic manipulator system may be computed as a by-product of the Newton-Euler dynamics formulation of the manipulator (Longman, R., Lindberg, R., & Zedd, M., 1987) which computes the reaction moment exerted at the manipulator mount point on the spacecraft as:

$$N_r = N_T + (p_{cm}^* - r_{c0} - s_0) \times F_T \qquad (3)$$

where

$$F_T = \sum_{i=1}^{n+1} F_{ci} = \sum_{i=1}^{n+1} m_i \dot{v}_{ci}$$

$$N_T = \sum_{i=1}^{n+1} N_{ci} = \sum_{i=1}^{n+1} I_i \dot{w}_i + w_i \times I_i w_i \qquad (4)$$

The values of $F_T$ and $N_T$ are computed as a by-product of the Newton-Euler dynamics formulation for the manipulator to ensure that $R_0 = I_3$. This provides the basis for stabilisation of the attitude of the spacecraft platform. The translation effect needs to be compensated for, and this can be done through a variation on the terrestrial Denavit-Hartenburg matrix formulation thus (Ellery, A., 2004b):

$$q = \begin{pmatrix} R & p^* \\ 0 & 1 \end{pmatrix} \tag{5}$$

where R=3x3 direction cosines matrix as for terrestrial manipulators

$$p^* = p_{cm}^* + \left(\frac{m_0}{m_T}\right)s_0 + \sum_{i=1}^{n} R_i \lambda_i - \left(\frac{m_{n+1}}{m_T}\right) R_{n+1} r_{n+1} \tag{6}$$

= inertial position of the end effector   $\lambda_i = \frac{1}{m_T}\left(\sum_{j=0}^{i} m_j l_i - m_i r_i\right)$ (7)

= kinematic-dynamic parameter of each link i

Similarly, the Jacobian may be given by $\bar{J} = \sum_{i=1}^{n}\sum_{k=1}^{i} \frac{\partial R_i}{\partial \theta_k}.\lambda_i$ providing the basis for resolved motion control algorithms (Ellery, A., 2004b) such as the computed torque controller and force control algorithms such as the hybrid position/force controller. This is applicable to any geometry of manipulator of n degrees of freedom, which is determined by the four Denavit-Hartenburg parameters. These results mean that terrestrial robotic control algorithms may be used with only minor modifications for the control of a space manipulator, easing the computational burden on space-rated processors for real-time control.

## 4. Implications

I have found that use of the above algorithm suggests that realistic servicer designs such as ATLAS require the use of control moment gyroscopes for spacecraft attitude control, particularly when implementing force control (Ellery, A., 2004a). The formulation presented above is readily extended to two or more manipulators allowing the implementation of onboard closed loop control of space manipulators mounted onto robotic servicer spacecraft. Robotic on-orbit servicing of spacecraft is achievable today and indeed, a number of space agencies are currently investigating this possibility in the near future.

The opportunity for OOS has important implications for the development of the space environment – in providing a fundamental part of space infrastructure, OOS represents the first tentative steps towards the development of a fully functional space-based manufacturing capability with material processing and assembly, and only when this is achieved, will space become a part of the everyday world.